\documentclass[10pt,twocolumn,letterpaper]{article}

\usepackage[pagenumbers]{cvpr} 

\usepackage{graphicx}
\usepackage{amsmath}
\usepackage{amssymb}
\usepackage{booktabs}
\usepackage{pifont}
\usepackage{multirow}
\usepackage{makecell}
\usepackage{verbatim}
\usepackage[flushleft]{threeparttable}
\usepackage{algorithm}
\usepackage{listings}
\usepackage{color}

\usepackage[pagebackref,breaklinks,colorlinks]{hyperref}

\usepackage[capitalize]{cleveref}
\crefname{section}{Sec.}{Secs.}
\Crefname{section}{Section}{Sections}
\Crefname{table}{Table}{Tables}
\crefname{table}{Tab.}{Tabs.}
\newcommand{\cmark}{\ding{51}}%
\newcommand{\xmark}{\ding{55}}%

\def\cvprPaperID{3650} 
\def\confName{CVPR}
\def\confYear{2023}

\begin{document}

\title{VoxelNeXt: Fully Sparse VoxelNet for 3D Object Detection and Tracking}

\author{Yukang Chen$^{1}$,
~~~
Jianhui Liu$^{2}$,~~~
Xiangyu Zhang$^{3}$,~~~
Xiaojuan Qi$^{2}$,~~~
Jiaya Jia$^{1}$
\\[0.2cm]
$^1$The Chinese University of Hong Kong~~
$^2$The University of Hong Kong~~
$^3$MEGVII Technology~~
}
\maketitle

\begin{abstract}
   3D object detectors usually rely on hand-crafted proxies, {\em e.g.}, anchors or centers, and translate well-studied 2D frameworks to 3D. Thus, sparse voxel features need to be densified and processed by dense prediction heads, which inevitably costs extra computation.
   In this paper, we instead propose VoxelNext for fully sparse 3D object detection. Our core insight is to predict objects directly based on sparse voxel features, without relying on hand-crafted proxies.
   Our strong sparse convolutional network VoxelNeXt detects and tracks 3D objects through voxel features entirely. It is an elegant and efficient framework, with no need for sparse-to-dense conversion or NMS post-processing. Our method achieves a better speed-accuracy trade-off than other mainframe detectors on the nuScenes dataset.
   For the first time, we show that a fully sparse voxel-based representation works decently for LIDAR 3D object detection and tracking. Extensive experiments on nuScenes, Waymo, and Argoverse2 benchmarks validate the effectiveness of our approach. Without bells and whistles, our model outperforms all existing LIDAR methods on the nuScenes tracking test benchmark. Code and models are available at \href{https://github.com/dvlab-research/VoxelNeXt}{github.com/dvlab-research/VoxelNeXt}.
   
\end{abstract}

\section{Introduction}
\label{sec:intro}
3D perception is a fundamental component in autonomous driving systems. 3D detection networks take sparse point clouds or voxels as input, and localize and categorize 3D objects. Recent 3D object detectors~\cite{pvrcnn,centerpoint,voxel-rcnn} usually apply sparse convolutional networks~(Sparse CNNs)~\cite{second} for feature extraction owing to its efficiency. Inspired by 2D object detection frameworks~\cite{fasterrcnn,centernet}, anchors~\cite{second, voxel-rcnn} or centers~\cite{centerpoint}, {\em i.e.}, dense point anchors in CenterPoint~\cite{centerpoint}, are commonly utilized for prediction. 
Both of them are hand-crafted and taken as intermediate proxies for 3D objects. 

Anchors and centers are designed for regular and grid-structured image data in the first place, and do not consider sparsity and irregularity of 3D data. To employ these proxy representations, the main stream of detectors~\cite{centerpoint, pvrcnn, voxel-rcnn} convert 3D sparse features to 2D dense features, so as to build a dense detection head for the ordered anchors or centers. Albeit useful, this dense head tradition leads to other limitations, including {\em inefficiency} and {\em complicated pipelines}, as explained below.
\begin{figure}[t]
\begin{center}
   \includegraphics[width=\linewidth]{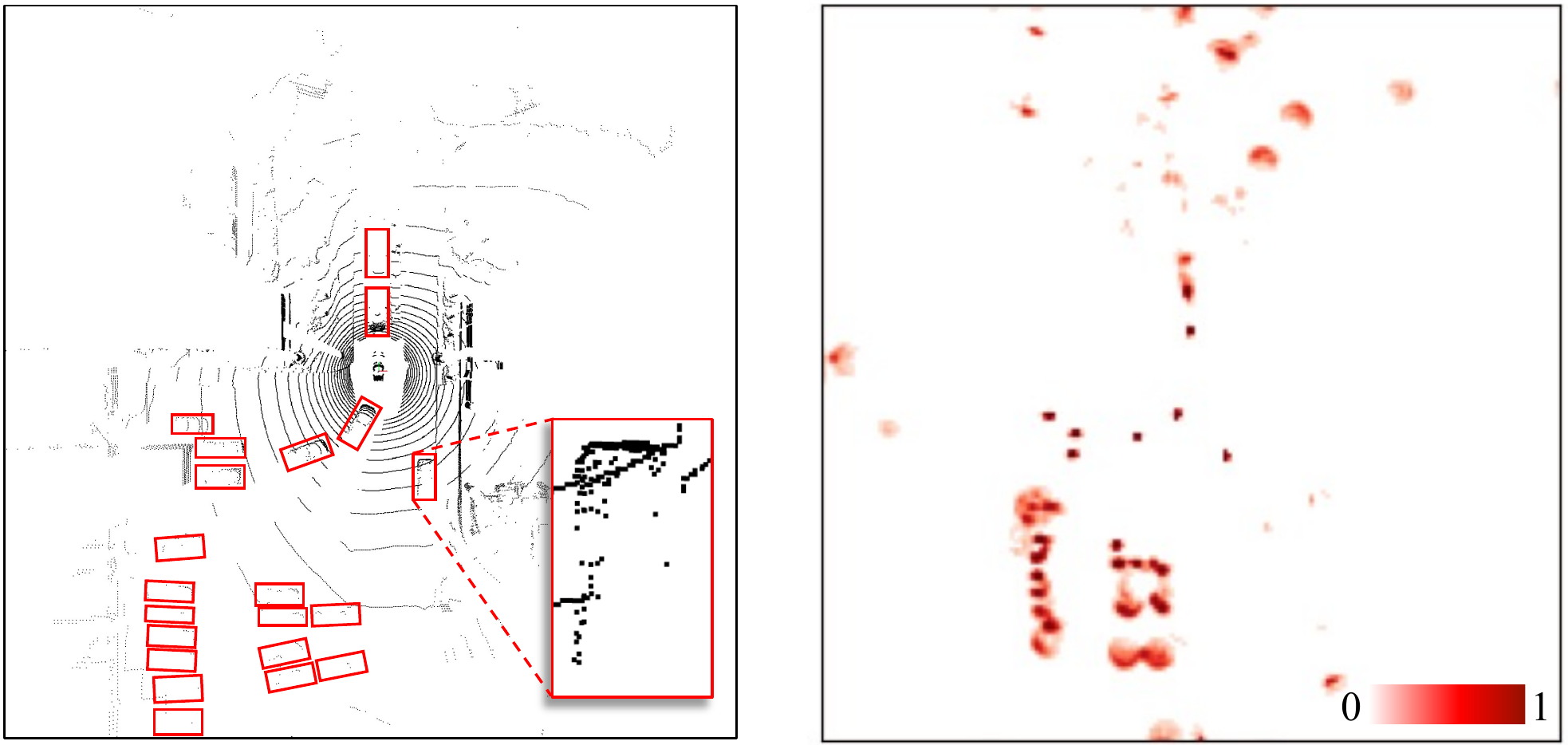}
   \caption{Visualization of input and heatmaps of CenterPoint in BEV for $Car$. Most values in the heatmaps are nearly zero, while the dense head computes over all BEV features, which is wasteful.}
   \label{fig:centerpoint-bev}
\end{center}
\end{figure}

\begin{figure*}[t]
\begin{center}
   \includegraphics[width=\linewidth]{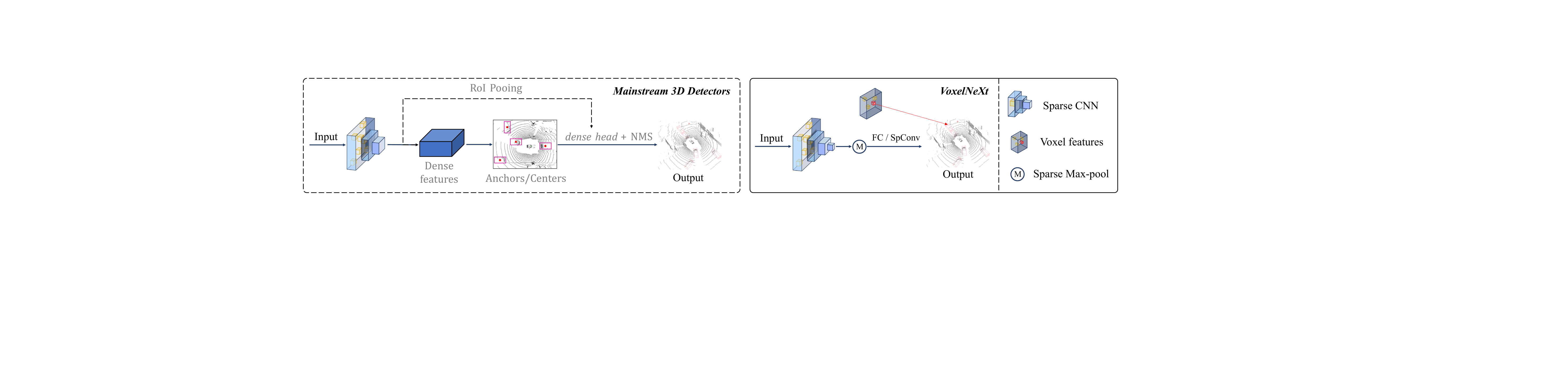}
   \caption{Pipelines of mainstream 3D object detectors and VoxelNeXt. These 3D detectors~\cite{centerpoint,pvrcnn,voxel-rcnn} rely on sparse-to-dense conversion, anchors/centers, and dense heads with NMS. RoI pooling is an option for two-stage detectors~\cite{voxel-rcnn,pvrcnn}. In contrast, VoxelNeXt is a fully sparse convolutional network, which predicts results directly upon voxel features, with either fully connected layers or sparse convolutions.}
   \label{fig:framework-comparison}
\end{center}
\end{figure*}

In Fig.~\ref{fig:centerpoint-bev}, we visualize the heatmap in CenterPoint~\cite{centerpoint}. It is clear that a large portion of space has nearly zero prediction scores. Due to inherent sparsity and many background points, only a small number of points have responses, {\em i.e.}, less than 1\% for {\em Car} class on average of nuScenes validation set. However, the dense prediction head computes over all positions in the feature map, as required by the dense convolution computation. They not only waste much computation, but also {\em complicate detection pipelines} with redundant predictions. It requires to use non-maximum suppression~(NMS) like post-processing to remove duplicate detections, preventing the detector from being elegant. These limitations motivate us to seek alternative sparse detection solutions.

In this paper, we instead propose {\em VoxelNeXt}. It is a simple, efficient, and post-processing-free 3D object detector. The core of our design is a voxel-to-object scheme, which directly predicts 3D objects from voxel features,  with a strong fully sparse convolutional network.
The key advantage is that our approach can get rid of anchor proxies, sparse-to-dense conversion, region proposal networks, and other complicate components. We illustrates the pipelines of mainstream 3D detectors and ours in Fig.~\ref{fig:framework-comparison}.

High inference {\em efficiency} is due to our {\em voxel-to-object} scheme avoiding dense feature maps. It predicts only upon sparse and necessary locations, as listed in Tab.~\ref{tab:computation-comparison} with comparison to CenterPoint~\cite{centerpoint}.
This representation also makes {\em VoxelNeXt} easily extended to {\em 3D tracking} with an offline tracker. Previous work~\cite{centerpoint} only tracks for the predicted object centers, which might involve prediction bias to its positions. In VoxelNeXt, the {\em query voxels}, {\em i.e.}, the voxels for box prediction, can also be tracked for association.

\begin{figure}[t]
\begin{center}
   \includegraphics[width=\linewidth]{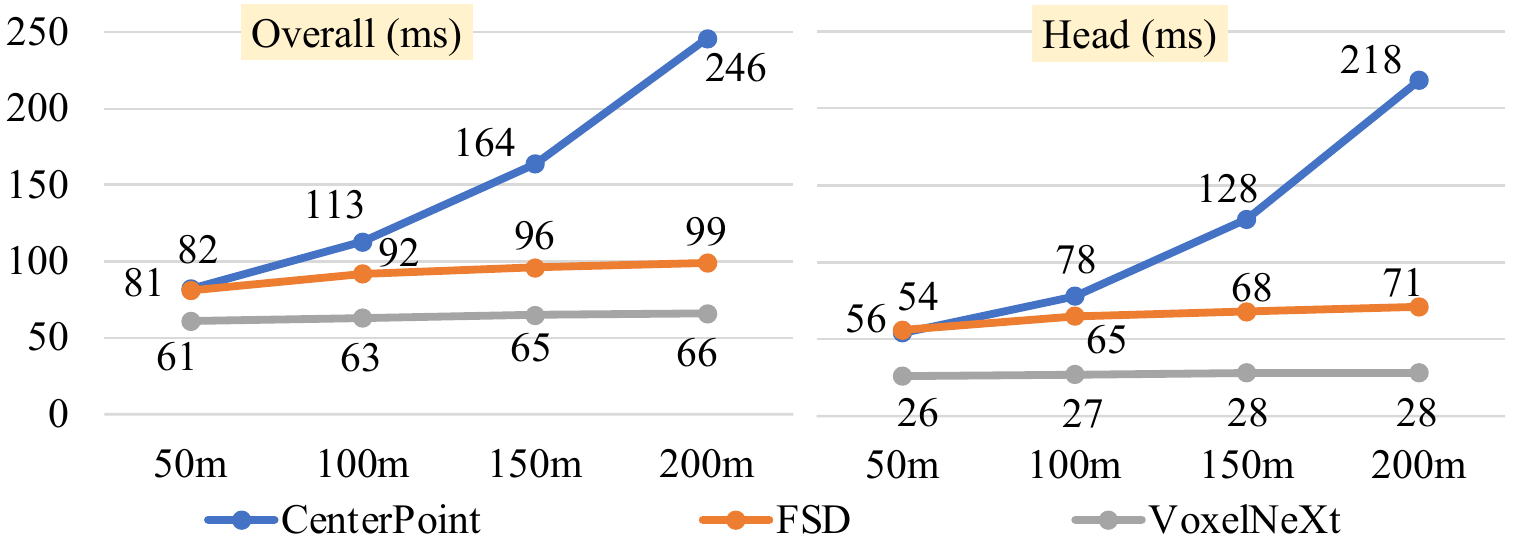}
   \caption{Latency on Argoverse2 and various perception ranges.}
   \label{fig:argo2-efficiency-comparison}
\end{center}
\end{figure}

Recently, FSD~\cite{fsd} exploits the fully sparse framework. Motivated by VoteNet~\cite{votenet}, it votes for object centers and resorts to iterative refinement.
Since 3D sparse data is generally scattered on object surfaces, this voting process inevitably introduces bias or error. Consequently, refinement, such as iterative group correction, is needed to ensure final accuracy. The system is complicated by its heavy belief in object centers. FSD~\cite{fsd} is promising at the large-range Argoverse2, while its efficiency is inferior to ours, as in Fig.~\ref{fig:argo2-efficiency-comparison}.

To demonstrate the effectiveness of {VoxelNeXt}, we evaluate our models on three large-scale benchmarks of nuScenes~\cite{nuscenes}, Waymo~\cite{waymo}, Argoverse2~\cite{argo2} datasets. VoxelNeXt achieves leading performance with high efficiency on 3D object detection on both these benchmarks. It also yields state-of-the-art performance on 3D tracking. Without bells and whistles, it ranks 1$^{st}$ among all LIDAR-only entries on the nuScenes tracking test split~\cite{nuscenes}.


\section{Related Work}
\label{sec:related_work}

\noindent
\textbf{LIDAR Detectors}
3D detectors usually work similar to their 2D counterparts, such as R-CNN series~\cite{voxel-rcnn, pyramid-rcnn, pvrcnn, graph-rcnn} and CenterPoint series~\cite{centerpoint,centernet,centertrack}. 3D detection distinguishes from the 2D task due to the sparsity of data distribution. But many approaches~\cite{voxelnet,second,centerpoint,voxel-rcnn} still seek 2D dense convolutional heads as a solution.

\begin{table}[t]
\begin{center}
\caption{Comparison with CenterPoint on nuScenes dataset. VoxelNeXt presents better performance with high efficiency.}
\resizebox{0.92\linewidth}{!}{
\begin{tabular}{|l|cc|cc|}
\hline
\multirow{2}{*}{\em Method} & \multirow{2}{*}{mAP} & \multirow{2}{*}{NDS} & \multicolumn{2}{c|}{FLOPs} \\
                        &                      &                      & Sparse CNN    & Head  \\ \hline
CenterPoint~\cite{centerpoint}             & 58.6                 & 66.2                 & 62.9 G  & 123.7 G \\
VoxelNeXt               & \textbf{60.0}    & \textbf{67.1}                 & \textbf{33.6 G} & \textbf{5.1 G} \\ \hline
\end{tabular}}
\label{tab:computation-comparison}
\end{center}
\end{table}
VoxelNet~\cite{voxelnet} uses PointNet~\cite{pointnet++} for voxel feature encoding and then applies dense region proposal network and head for prediction. SECOND~\cite{second} improves VoxelNet by efficient sparse convolutions with the dense anchor-based head. Other state-of-the-art methods, including PV-RCNN~\cite{pvrcnn}, Voxel R-CNN~\cite{voxel-rcnn}, and VoTr~\cite{voxeltransformer}, still keep the sparse-to-dense scheme to enlarge the receptive field. 

Motivated by 2D CenterNet~\cite{centernet}, CenterPoint~\cite{centerpoint} is applied to 3D detection and tracking. It converts the sparse output of a backbone network into a map-view dense feature map and predicts a dense heatmap of the center locations of objects, based on the dense feature. This dense center-based prediction has been adopted by several dense-head approaches~\cite{bevfusion-mit,bevfusion-ali}. In this paper, we take a new direction and surprisingly show that a simple and strong sparse CNN is sufficient for direct prediction. The notable finding is that the dense head is not always necessary.

\vspace{0.5em}
\noindent
\textbf{Sparse Detectors}
Methods of~\cite{fsd,rsn,swformer} avoid dense detection heads and instead introduce other complicated pipelines. RSN~\cite{rsn} performs foreground segmentation on range images and then detects 3D objects on the remained sparse data. SWFormer~\cite{swformer} proposes a sparse transformer with delicate window splitting and multiple heads with feature pyramids. Motivated by VoteNet~\cite{votenet}, FSD~\cite{fsd}
use point clustering and group correction to solve the issue of center feature missing. These detectors conduct sparse prediction, but complicate detection pipelines in different ways. In our work, this center-missing issue can also be simply skipped through sparse networks that have large receptive fields. We make minimal adaptations to commonly-used sparse CNNs to realize fully sparse detectors.

\vspace{0.5em}
\noindent
\textbf{Sparse Convolutional Networks}
Sparse CNNs become mainframe backbone networks in 3D deep learning~\cite{pvrcnn, chu2021icm, chu2022twist, jiang2021guided} for its efficiency. It is common wisdom that its representation ability is limited for prediction. To remedy it, 3D detectors of~\cite{second,pvrcnn,voxel-rcnn,infofocus} rely on dense convolutional heads for feature enhancement. Recent methods~\cite{focalsconv,spatial-pruned-conv} make convolutional modifications upon sparse CNNs. Approaches of~\cite{voxeltransformer,voxelset} even substitute it with transformers for large receptive fields. Contrary to all these solutions, we demonstrate that the insufficient receptive field bottleneck can be simply addressed by additional down-sampling layers without any other complicated design.

\begin{figure}[t]
\begin{center}
   \includegraphics[width=\linewidth]{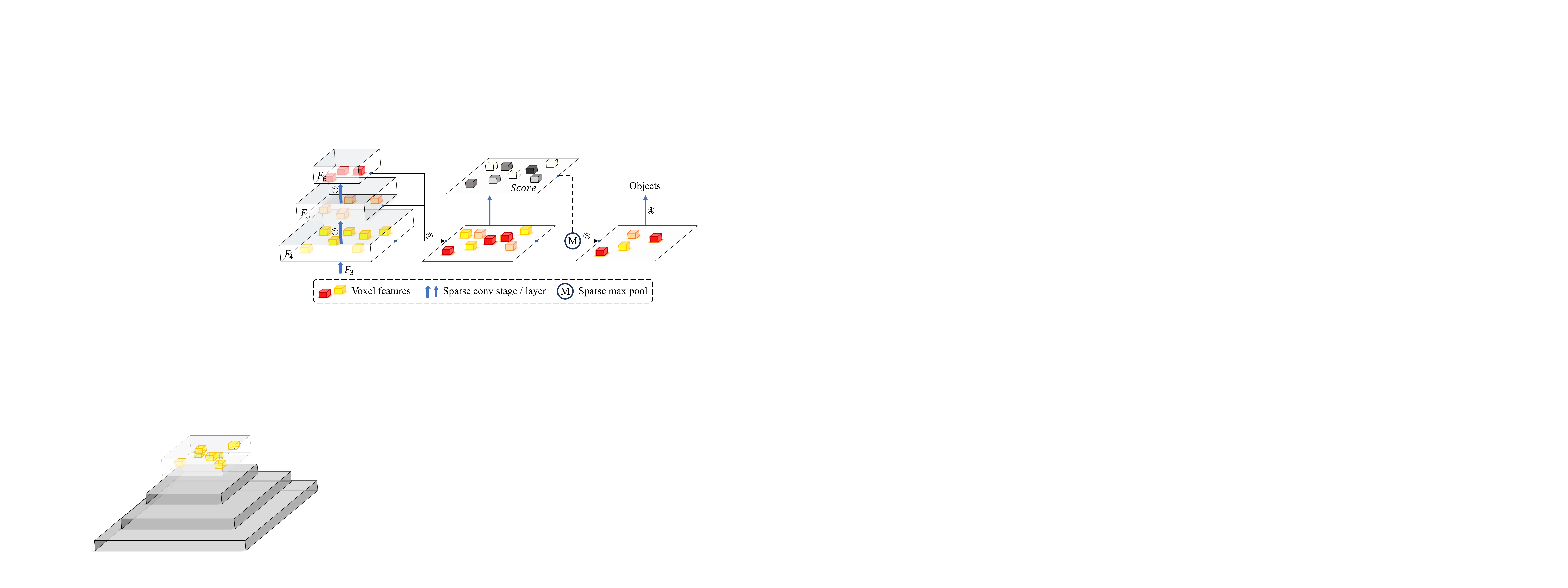}
   \caption{Detailed structure of VoxelNeXt framework. Circled numbers in the figure correspond to the paragraphs in Sections~\ref{sec:backbone-adaptation} and \ref{sec:sparse-prediction-head}. 
   1 - Additional down-samplings. 2 - Sparse height compression. 3 - Voxel selection. 4 - Box regression. 
   We omit the generation of $F_1$, $F_2$, and $F_3$ here for the simplicity sake.}
   \label{fig:voxelnext-details}
\end{center}
\end{figure}

\vspace{0.5em}
\noindent
\textbf{3D Object Tracking}
3D object tracking models tracklets of multiple objects along multi-frame LIDAR. Most previous methods~\cite{probabilistic-tracking,cbmot,ab3dmot} directly use the Kalman filter upon detection results, such as AB3DMOT~\cite{ab3dmot}. CenterPoint~\cite{centerpoint} predicts the velocities to associate object centers through multiple frames, following CenterTrack~\cite{centertrack}. In this paper, we include query voxels for association, which effectively relieve the prediction bias of object centers.

\section{Fully Sparse Voxel-based Network}
\label{sec:voxelnext}
Point clouds or voxels are irregularly distributed and usually scattered at the surface of 3D objects, not at the center or inside. This motivates us to study along a new direction to {\em predict 3D boxes directly based on the voxels} instead of the hand-crafted anchors or centers.

To this end, we aim for {\em minimal modification} to adapt a plain 3D sparse CNN network to the direct-voxel prediction. In the following, we introduce the backbone adaptation (Section~\ref{sec:backbone-adaptation}), the sparse head design (Section~\ref{sec:sparse-prediction-head}), and the extension to 3D object tracking (Section~\ref{sec:3d-tracking}).

\begin{figure}[t]
\begin{center}
   \includegraphics[width=\linewidth]{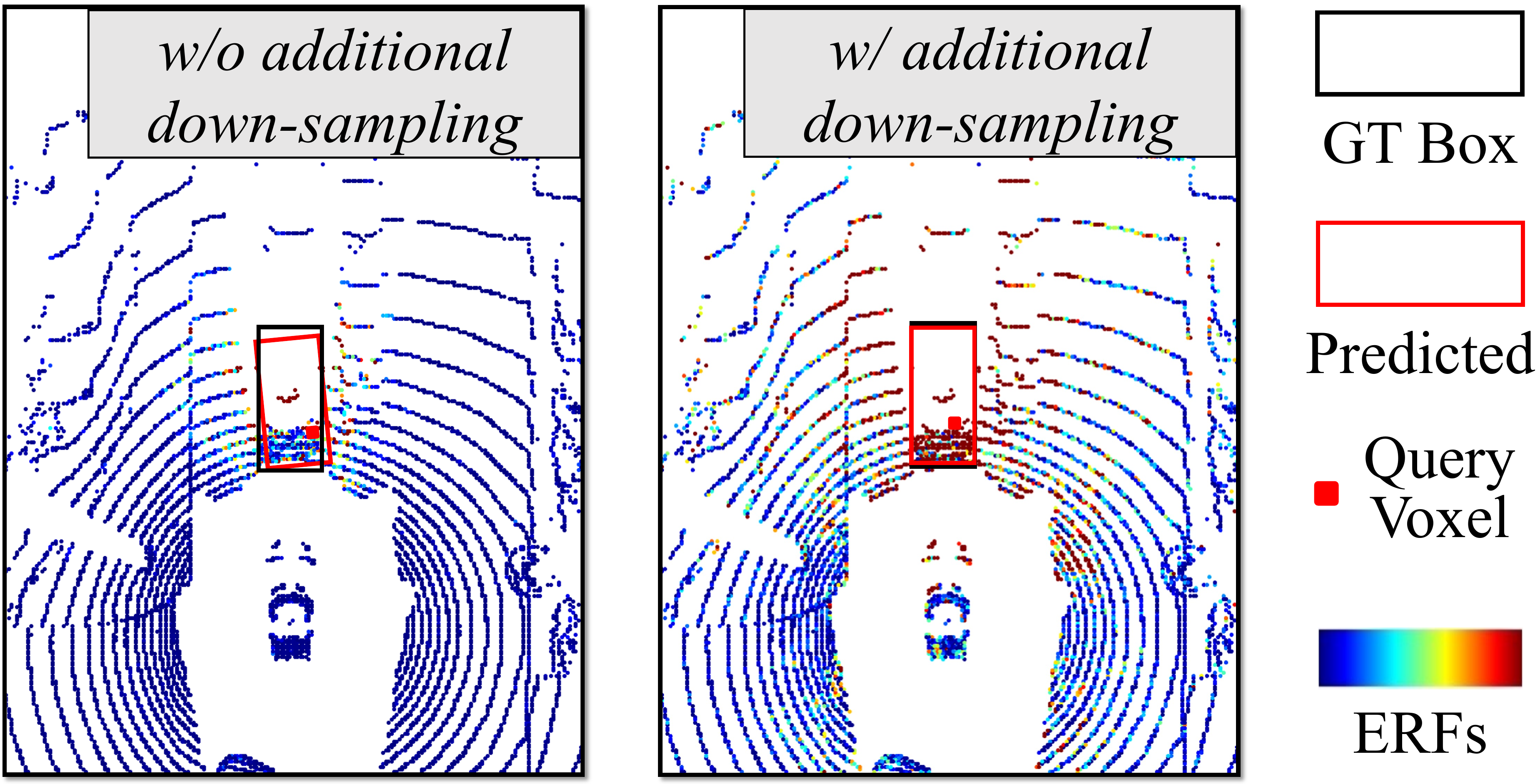}
   \caption{Effects of additional down-sampling layers on effective receptive fields~(ERFs) and the predicted boxes.}
   \label{fig:effective-receptive-fields}
\end{center}
\end{figure}
\subsection{Sparse CNN Backbone Adaptation}
\label{sec:backbone-adaptation}
\noindent
\textbf{Additional Down-sampling}
Strong feature representation with sufficient receptive fields is a must to ensure direct and correct prediction upon sparse voxel features.
Although the plain sparse CNN backbone network has been widely used in 3D object detectors~\cite{pvrcnn,voxel-rcnn, centerpoint}, recent work presents its weakness and proposes various methods to enhance the sparse backbone using, {\em e.g.}, well-designed convolution~\cite{focal-sparse-conv}, large kernels~\cite{largekernel3d}, and transformers~\cite{voxeltransformer, lai2022stratified, lai2023spherical}.

Unlike all these approaches, we make as little as possible modification to accomplish this, only using additional down-sampling layers. By default, the plain sparse CNN backbone network has 4 stages, with the feature strides \{1, 2, 4, 8\}. We name the output sparse features \{$F_1, F_2, F_3, F_4$\} respectively. This setting is incapable of direct prediction, especially for large objects. To enhance its ability, we simply include two additional down-sampling layers to obtain features with strides \{16, 32\} for \{$F_5, F_6$\}. This small change directly imposes notable effects to enlarge receptive fields. We combine the sparse features from the last three stages \{$F_4, F_5, F_6$\} to $F_c$. Their spatial resolutions are all aligned to $F_4$. For stage $i$, $F_i$ is a set of individual features $f_p$. $p\in P_i$ is a position in 3D space, with the coordinate ($x_p$, $y_p$, $z_p$). This process is shown in Fig.~\ref{fig:voxelnext-details}. It is noteworthy that this simple sparse concatenation requires no other parameterized layers. 
Sparse features $F_c$ and their positions $P_c$
are obtained as
\begin{equation}
\label{eq:combine}
\begin{aligned}
&{F}_{c}=F_4\cup(F_5\cup F_6),\qquad \\
&P_6' = \{(x_p \times 2^2, \, y_p \times 2^2, \, z_p \times 2^2) \,|\, p\in P_6\} \\
&P_5' = \{(x_p \times 2^1, \, y_p \times 2^1, \, z_p \times 2^1) \,|\, p\in P_5\} \\
&{P}_{c}=P_4\cup(P_5'\cup P_6').\qquad \\
\end{aligned}
\end{equation}

We visualize the effective receptive fields~(ERFs) in Fig.~\ref{fig:effective-receptive-fields}. With additional down-sampling layers, ERFs are larger and the predicted box is more accurate.
It is effective enough and costs little extra computation, as in Tab.~\ref{tab:nuscenes-downsamples-ablation}. Thus, we use this simple design as the backbone network.
\begin{figure}[t]
\begin{center}
   \includegraphics[width=\linewidth]{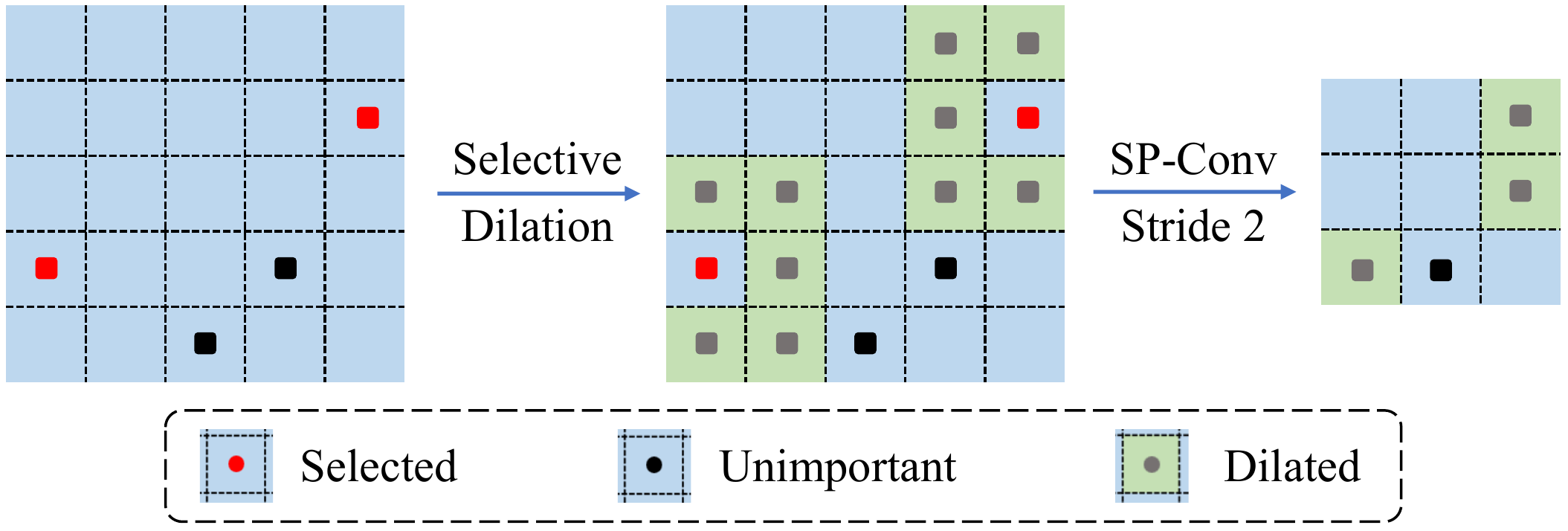}
   \caption{Spatially voxel pruning. In sparse CNN backbone, down-sampling layers commonly dilate all voxels to the kernel shape, before down-sampling. Different from these approaches, we only dilate selected voxels that have high feature magnitudes to maintain high efficiency.}
   \label{fig:selective-down-sampling}
\end{center}
\end{figure}
\begin{figure}[t]
\begin{center}
   \includegraphics[width=\linewidth]{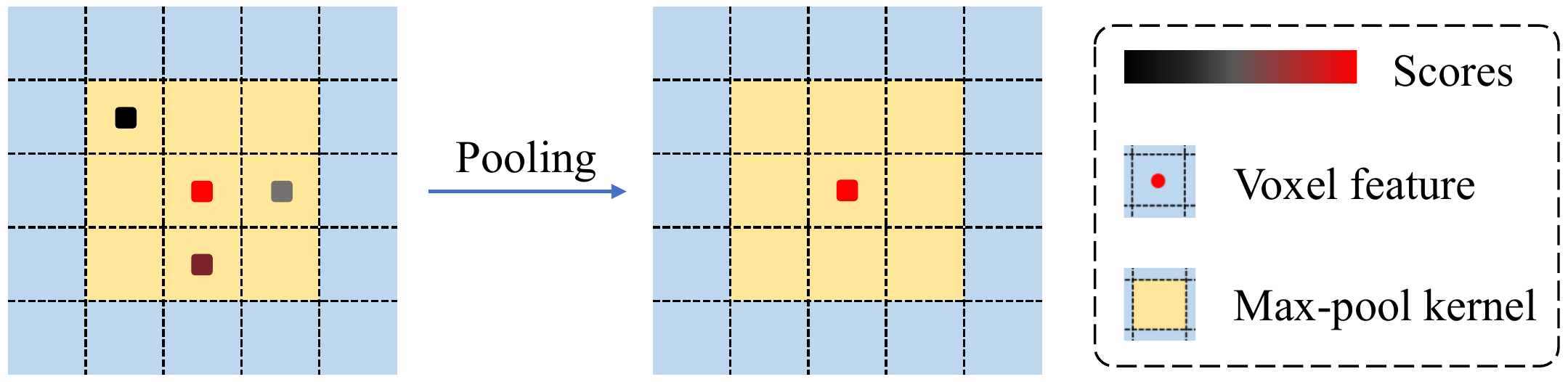}
   \caption{Sparse max pooling layer. Similarly to submanifold sparse convolution~\cite{submanifold-sparse-conv-v2}, it only operates on non-empty positions. It removes non-maximum voxels in local space.}
   \label{fig:sparse-max-pooling}
\end{center}
\end{figure}

\vspace{0.5em}
\noindent
\textbf{Sparse Height Compression}
3D object detectors of~\cite{centerpoint, pvrcnn, voxel-rcnn} compress 3D voxel features into dense 2D maps by converting sparse features to dense ones and then combining depth (along $z$ axis) into the channel dimension. These operations cost footprint memory and computation.

In VoxelNet, we find that 2D sparse features are efficient for prediction. Height compression in VoxelNeXt is fully sparse. We simply put all voxels onto the ground and sum up features in the same positions. It costs no more than 1ms. We find that prediction upon the compressed 2D sparse features cost less than using 3D ones, as shown in Tab.~\ref{tab:ablation-sparse-bev-compression}.The compressed sparse features $\bar{F_c}$ and their positions $\bar{P_c}$ are obtained as:
\begin{equation}
\label{eq:hight-compress}
\begin{aligned}
&\bar{P_c} = \{(x_p, \, y_p) \,|\, p\in P_c\} \\
&\bar{F_c} = \{\sum_{p\in S_{\bar{p}}} f_p ,|\, \bar{p}\in \bar{P_c}\}
\end{aligned}
\end{equation}
where $S_{\bar{p}} = \{p \, | \, x_p=x_{\bar{p}},\, y_p=y_{\bar{p}},\, p\in P_c\}$, containing voxels that are put onto the same 2D position $\bar{p}$.

\vspace{0.5em}
\noindent
\textbf{Spatially Voxel Pruning}
Our network is completely based on voxels. It is common that 3D scenes contain a large number of background points that is redundant and have little benefit for prediction. We gradually prune irrelevant voxels along down-sampling layers. Following SPS-Conv~\cite{spatial-pruned-conv}, we suppress the dilation of voxels with small feature magnitudes, as shown in Fig.~\ref{fig:selective-down-sampling}. Taking the suppression ratio as 0.5, we only dilate the voxels whose feature magnitudes $|f_p|$ (averaged on the channel dimension) rank top half of all voxels. The voxel pruning largely saves computation without compromising performance as indicated in Tab.~\ref{tab:nuscenes-sprs-ratio-ablation}. 

\begin{figure}[t]
\begin{center}
   \includegraphics[width=\linewidth]{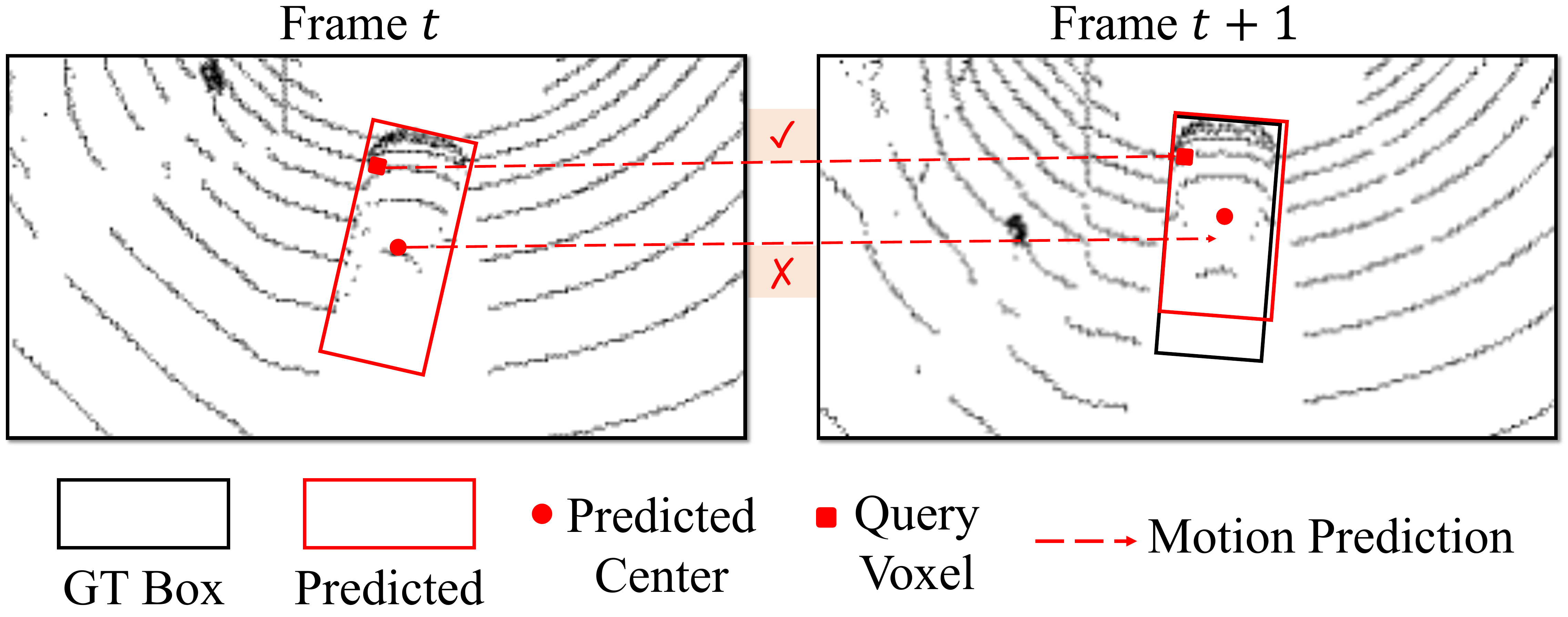}
   \caption{Visualization of voxel association. The predicted object centers are conventionally used for tracking. We additionally associate query voxels in case that the predicted centers are inaccurate.}
   \label{fig:voxel-association}
\end{center}
\end{figure}
\begin{figure}[t]
\begin{center}
   \includegraphics[width=0.96\linewidth]{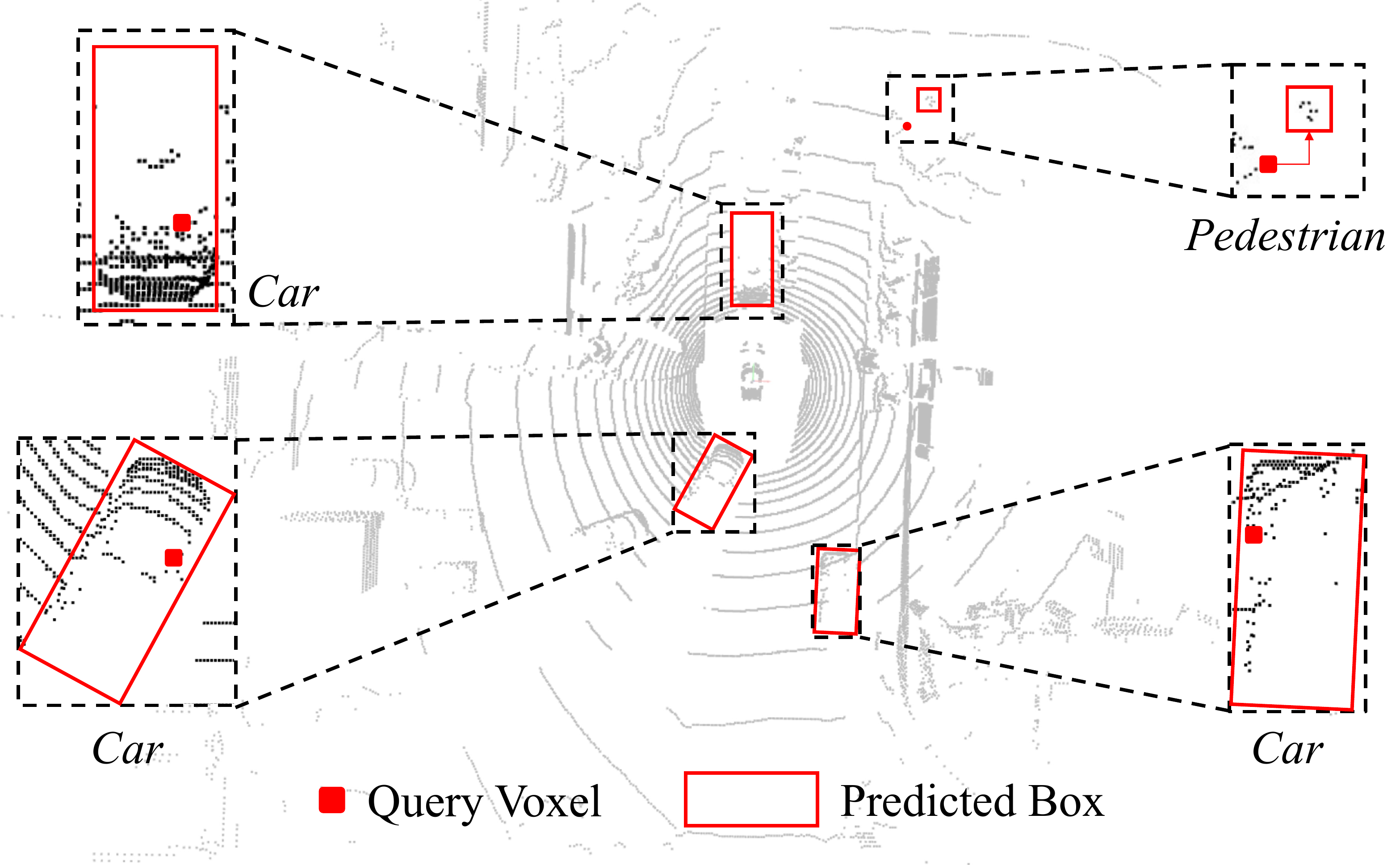}
   \caption{Visualization on the predicted boxes and their query voxels. For the {\em Car} objects, query voxels are inside and usually near the boundaries. For the pedestrian consisting of limited voxels, its query voxel is outside. More visualizations are in the appendix.}
   \label{fig:voxels-and-points}
\end{center}
\end{figure}


\subsection{Sparse Prediction Head}
\label{sec:sparse-prediction-head}

\noindent
\textbf{Voxel Selection}
Figure~\ref{fig:voxelnext-details} shows the detailed framework of the VoxelNeXt model. Instead of relying on the dense feature map $\mathbf{M}$, we directly predict objects based on the sparse output of the 3D CNN backbone network $\mathbf{V}\in \mathbb{R}^{N\times F}$. We first predict the scores of voxels for $K$ classes, $\mathbf{s}\in \mathbb{R}^{N\times K}$. During training, we assign the voxel nearest to each annotated bounding box center as a positive sample. We use a focal loss~\cite{focalloss} for supervision. 
We note the fact that during inference {\em query voxels are commonly not at the object center}.
They are even not necessarily inside the bounding boxes, {\em e.g.}, for pedestrian in Fig.~\ref{fig:voxels-and-points}. We count the distribution of query voxels in Tab.~\ref{tab:ratio-voxel-inside} on nuScenes validation set. 

\begin{table*}[t]
\begin{center}
\caption{Results of a pilot study on nuScenes validation split, for the strides of fully sparse voxel-based prediction. Latency is evaluated on a single GPU. For $D_3$, the arrows indicate the change based on CenterPoint. For others, the arrows indicate the change based on $D_3$.}
\resizebox{\linewidth}{!}{
\begin{tabular}{|l|c|c|cc|cccccccccc|}
\hline
{\em Method} & {\em Strides} & Latency  &  mAP  & NDS   & Car                  & Truck                & Bus                  & Trailer              & C.V.                   & Ped                  & Mot                  & Byc                  & T.C.                   & Bar               \\ \hline
CenterPoint & \{2, 4, 8\} & 96 ms &55.6 & 63.2  & 83.5 &
54.9 & 67.5 & 30.6 & 16.3 & 83.3 & 52.7 & 34.5 & 65.6 & 66.5 \\ \hline \hline
$D_{3}$ & \{2, 4, 8\} & 56 ms & 46.7\textcolor{red}{$_{\downarrow 8.9}$} & 56.2  & 75.3 & 41.3 & 38.3 & 10.5 & 14.9 & 82.0 & 47.7 & 28.3 & 63.6 & 64.2 \\
$D_{3}^{5\times 5 \times 5}$ & \{2, 4, 8\} & 225 ms  & 51.6\textcolor[rgb]{0,0.5,0}{$_{\uparrow 4.9}$} & 60.4 &  80.0 & 49.2 & 56.8 & 16.8 & 16.5 & 83.5 & 50.2 & 30.9 & 64.8 & 67.7 \\
$D_{4}$  & \{2, 4, 8, 16\} & 62 ms & 52.3\textcolor[rgb]{0,0.5,0}{$_{\uparrow 5.6}$} & 61.2  & 80.0 & 50.0 & 61.2 & 23.1 & 16.9 & 82.5 & 49.0 & 31.8 & 63.9 & 64.8 \\
$D_{5}$ & \{2, 4, 8, 16, 32\} & 66 ms & \textbf{56.5}\textcolor[rgb]{0,0.5,0}{$_{\uparrow 9.5}$} & \textbf{64.5}  & 83.0 & 54.0 & 67.4 & 32.9 & 20.0 & 84.1 & 52.7 & 35.7 & 66.6 & 65.3 \\
\hline
\end{tabular}
}
\label{tab:nuscenes-downsamples-ablation}
\end{center}
\end{table*}
\begin{table}[t]
\begin{center}
\caption{Effects of spatial pruning ratios. A larger pruning ratio means that fewer voxels remain in the sparse CNN backbone.}
\begin{tabular}{|l|cccccc|}
\hline
 {\em Ratio}   & - & 0.1 & 0.3 & 0.5 & 0.7 & 0.9  \\ \hline
 FLOPs (G) & 83.8 & 79.6 & 60.1 & 33.6 & 19.8 & 7.6 \\
 mAP & 56.5 & 56.5 & 56.4 & 56.2 & 53.7 & 45.1 \\
 NDS & 64.5 & 64.5 & 64.3 & 64.3 & 62.1 & 56.0 \\ \hline
\end{tabular}
\label{tab:nuscenes-sprs-ratio-ablation}
\end{center}
\end{table}
\begin{table}[t]
\begin{center}
\caption{Effects of spatial pruning on various layers. We use it on the first 3 down-sampling layers by default.}
\resizebox{\linewidth}{!}{
\begin{tabular}{|l|cccccc|}
\hline
{\em Stages} & - & 1 & 2 & 3 & 4 & 5 \\ \hline
FLOPs (G) & 83.8 & 65.0 & 45.9 & 33.6 & 29.1 & 27.9 \\
mAP & 56.5 & 56.5 & 56.4 & 56.2 & 54.2 & 53.7 \\ 
NDS & 64.5 & 64.5 & 64.4 & 64.3 & 62.5 & 62.0 \\ 
\hline
\end{tabular}}
\label{tab:voxel-pruning-layers}
\end{center}
\end{table}

During inference, we avoid NMS post-processing by using sparse max pooling, as features are sparse enough. Similar to submanifold sparse convolution~\cite{submanifold-sparse-conv-v2}, it only operates on non-empty positions. This is based on the predicted scores $\mathbf{s}$ and conducted individually for each class. We adopt sparse max pooling to select voxels with spatially local maximums.  The removed voxels will be excluded in box prediction, which saves the computation of head.

\vspace{0.5em}
\noindent
\textbf{Box Regression}
Bounding boxes are directly regressed from the positive or selected sparse voxel features $\mathbf{v}\in \mathbb{R}^{n\times F}$. Following the protocol in CenterPoint~\cite{centerpoint}, we regress the location $(\Delta x, \Delta y)\in \mathbb{R}^2$, height $h\in \mathbb{R}$, 3D size $s \in \mathbb{R}^3$, and rotation angle $(sin(\alpha), cos(\alpha)) \in \mathbb{R}^2$. For the nuScenes dataset or tracking, we regress the velocity $v \in \mathbb{R}^2$ by task definition. These predictions are supervised under the L1 loss function during training. For Waymo dataset, we also predict the IoU and train with IoU loss for performance enhancement~\cite{afdetv2}. We simply use fully connected layer or $3\times 3$ submanifold sparse convolutional layers with kernel size 3 for prediction, without other complicate designs. We find that the $3\times 3$ sparse convolutions generate better results than fully connected layers, with limited burden, as in Tab.~\ref{tab:head-sparse-conv}.

\subsection{3D Tracking}
\label{sec:3d-tracking}
Our framework is naturally extended to 3D tracking. CenterPoint~\cite{centerpoint} tracks the predicted object centers via a two-dimensional velocity $v \in \mathbb{R}^2$, which is also supervised by L1 loss. We extend this design into VoxelNeXt. Our solution is to use {\em voxel association} to include more tracklets that match the positions of query voxels.

As shown in Fig.~\ref{fig:voxel-association}, we record the position of voxel that is used to predict each box. Similar to the center association, we compute the L2 distance for matching. The query positions are picked by tracking back their index to original input voxels, instead of stride-8 positions.  The tracked voxels exist in input data, which has less bias than the predicted centers. Also, the query voxels between adjacent frames share similar relative positions to boxes. We empirically show that voxel association improves tracking in Tab.~\ref{tab:nuscenes-voxel-tracking}.

\section{Experiments}
\label{sec:experiments}

\begin{table}[t]
\begin{center}
\caption{Ablations on 2D or 3D sparse CNN in VoxelNeXt. sparse height Compression is used to connect 3D backbone and 2D head.}
\resizebox{\linewidth}{!}{
\begin{tabular}{|l|cc|c|cc|}
\hline
 {\em Method} & {\small Backbone} & {\small Head}  & Latency &  mAP    & NDS  \\ \hline
 - & 3D & 3D  & 92 ms & 56.3 & 63.4 \\
 {\small VoxelNeXt} & 3D & 2D  & 66 ms & \textbf{56.2} & \textbf{64.3} \\
 {\small VoxelNeXt-2D} & 2D & 2D  & \textbf{61 ms} & 53.4 & 62.6 \\
\hline
\end{tabular}
}
\label{tab:ablation-sparse-bev-compression}
\end{center}
\end{table}
\begin{table}[t]
\begin{center}
\caption{Effects of the layer type in the sparse prediction head. $1\times 1$ submanifold sparse convolution~\cite{submanifold-sparse-conv-v2} is the fully connected.}
\begin{tabular}{|l|c|cc|}
\hline
{Head kernel size} &  Head latency   &   mAP    & NDS  \\ \hline
$1\times 1$ (FC) & 30 ms & 56.2 & 64.3 \\
$3\times 3$ (SpConv) & 35 ms & 56.8 & 64.5 \\
\hline
\end{tabular}
\label{tab:head-sparse-conv}
\end{center}
\end{table}

\begin{table*}[t]
\begin{center}
\caption{Ratios of relative positions of query voxels to the boxes predicted from them. We only take high-quality predicted boxes (IoU with ground-truth boxes $>$ 0.7 and with matched predicted labels) into consideration. According to the relative positions to their predicted boxes, we split voxels into 3 types of {\em near center}, {\em near boundary}, and {\em outside box}. Overall, most voxels are inside but not near center.}
\resizebox{\linewidth}{!}{
\begin{tabular}{|l|c|cccccccccc|}
\hline
{\em Class} & Mean                             & Car                              & Truck                            & Bus                              & Trailer                          & C.V.                             & Ped                              & Mot                              & Byc                              & T.C.                             & Bar                              \\ \hline
Near center                 & 9.9\%                            & 10.3\%                           & 5.6\%                            & 15.2\%                           & 1.2\%                            & 16.3\%                           & 12.5\%                           & 19.6\%                           & 13.1\%                           & 10.8\%                           & 17.8\%                           \\
Near boundary                & \textbf{72.8\%} & \textbf{84.3\%} & 39.2\%                           & \textbf{58.8\%} & \textbf{84.6\%} & \textbf{51.8\%} & 42.3\%                           & \textbf{66.5\%} & \textbf{54.7\%} & 39.7\%                           & \textbf{58.7\%} \\
Outside box                & 17.3\%                           & 5.4\%                            & \textbf{55.3\%} & 26.0\%                           & 14.2\%                           & 31.9\%                           & \textbf{45.2\%} & 13.9\%                           & 32.2\%                           & \textbf{49.6\%} & 23.5\%                           \\ \hline
\end{tabular}}
\label{tab:ratio-voxel-inside}
\end{center}
\end{table*}

\begin{table}[t]
\begin{center}
\caption{Comparison to the representative dense-head method Centerpoint~\cite{centerpoint}. ATE, ASE, AOE, AVE, and AAE denote the errors of location, size, orientation, velocity, and attribute.}
\resizebox{\linewidth}{!}{
\begin{tabular}{|c|cc|ccccc|}
\hline
{\em Method} &   mAP   & NDS  & ATE & ASE & AOE & AVE & AAE \\ \hline
 CenterPoint & 55.6 & 63.5 & \textbf{29.7} & 25.7 & 44.5 & 24.5 & \textbf{18.8} \\
\multirow{2}{*}{VoxelNeXt} & \textbf{56.5} & \textbf{64.5} & 29.9 & \textbf{25.4} & \textbf{39.6} & \textbf{23.2} & 19.0 \\
  & $\uparrow$0.9 & $\uparrow$1.0 & $\uparrow$0.2 & $\downarrow$0.3 & $\downarrow$\textbf{4.9} & $\downarrow$1.3 & $\uparrow$0.2 \\
\hline
\end{tabular}}
\label{tab:nuscenes-error-analysis}
\end{center}
\end{table}
\begin{table}[t]
\begin{center}
\caption{Efficiency statistics on sparse CNN backbone. The computations of Stage 5\&6 are limited by their small voxel numbers.}
\resizebox{\linewidth}{!}{
\begin{tabular}{|l|cccccc|}
\hline
{\em Stage} &  1   &   2    & 3  & 4 & 5 & 6 \\ \hline
Channel & 16  & 32 & 64 & 128 & 128 & 128 \\ 
Voxels (K) & 82.6  & 46.7  & 18.2  & 6.4 & 3.0 & 1.3\\ 
FLOPs (G) & 1.1  & 4.7 & 8.2 & 11.7 & 6.1 & 2.8 \\ 
Latency (ms) & 4 & 5 & 6 & 7 & 6 & 3 \\ 
\hline
\end{tabular}}
\label{tab:nuscenes-efficiency-stages}
\end{center}
\end{table}
\begin{table}[t]
\begin{center}
\caption{Effects of sparse max-pool and NMS post-processing. The max-pool follows the submanifold sparse convolution pattern.}
\begin{tabular}{|c|c|cc|}
\hline
{\em Max-pool} &  {\em NMS}   &   mAP    & NDS  \\ \hline
 \xmark & \xmark & 33.0 & 51.0 \\
 \xmark & \cmark & 56.0 & 64.2 \\
\cmark & \xmark & 56.2 & 64.3 \\
\cmark & \cmark & 56.2 & 63.3 \\
\hline
\end{tabular}
\label{tab:nuscenes-maxpool-ablation}
\end{center}
\end{table}
\begin{table}[t]
\begin{center}
\caption{Voxel association on nuScenes tracking validation set.}
\begin{tabular}{|c|c|ccc|}
\hline
\makecell[c]{\em + Voxel\\\em association}      &   AMOTA    & AMOTP & MOTA & IDS \\ \hline
\xmark & 69.1 & 61.6 & 59.3 & 643 \\
\cmark   & \textbf{70.2} & 64.0 & 61.5 & 729 \\
\hline
\end{tabular}
\label{tab:nuscenes-voxel-tracking}
\end{center}
\end{table}

\begin{table*}[t]
\begin{center}
\caption{Performance of 3D object detection methods on nuScenes test set. $^{\dagger}$ means the method that uses double-flip testing. All models listed take LIDAR data as input without image fusion or any model ensemble.}
\resizebox{\linewidth}{!}{
\begin{tabular}{|l|ccc|cccccccccc|}
\hline
                   {\em Method}          & mAP  & NDS  & Latency & Car                  & Truck                & Bus                  & Trailer              & C.V.                   & Ped                  & Mot                  & Byc                  & T.C.                   & Bar                   \\ \hline \hline
PointPillars~\cite{pointpillars}                    & 30.5 & 45.3 & 31 ms & 68.4 & 23.0 & 28.2 & 23.4 & 4.1 & 59.7 & 27.4 & 1.1 & 30.8 & 38.9 \\ 
3DSSD~\cite{3dssd}                        & 42.6 & 56.4 &  -  &  81.2 & 47.2 & 61.4 & 30.5 & 12.6 & 70.2 & 36.0 & 8.6 & 31.1 & 47.9 \\ 
CBGS~\cite{cbgs}                        & 52.8 & 63.3 &  80 ms & 81.1 & 48.5 & 54.9 & 42.9 & 10.5 & 80.1 & 51.5 & 22.3 & 70.9 & 65.7 \\ 
CenterPoint~\cite{centerpoint}                        & 58.0 & 65.5 & 96 ms & 84.6 & 51.0 & 60.2 & 53.2 & 17.5 & 83.4 & 53.7 & 28.7 & 76.7 & 70.9 \\ 
CVCNET~\cite{cvcnet}                 & 58.2 & 66.6 & 122 ms &  82.6 & 49.5 & 59.4 & 51.1 & 16.2 & 83.0 & 61.8 & 38.8 & 69.7 & 69.7 \\ 
HotSpotNet~\cite{hotspotnet}                  & 59.3 & 66.0  & - & 83.1 & 50.9 & 56.4 & 53.3 & 23.0 & 81.3 & 63.5 & 36.6 & 73.0 & 71.6 \\
AFDetV2~\cite{afdetv2}  & 62.4 & 68.5 &  - & 86.3 & 54.2 & 62.5 & 58.9 & 26.7 & 85.8 & 63.8 & 34.3 & 80.1 & 71.0 \\ 
Focals Conv~\cite{focal-sparse-conv}  & 63.8 & 70.0 &  138 ms  & 86.7 & 56.3 & 67.7 & 59.5 & 23.8 & 87.5 & 64.5 & 36.3 & 81.4 & 74.1 \\ 
VISTA~\cite{vista}$^{\dagger}$  & 63.0 & 69.8  & 94 ms & 84.4 & 55.1 & 63.7 & 54.2 & 25.1 & 82.8 & 70.0 & 45.4 & 78.5 & 71.4 \\ 
UVTR-L~\cite{uvtr}$^{\dagger}$  & 63.9 & 69.7 &  132 ms &  86.3 & 52.2 & 62.8 & 59.7 & 33.7 & 84.5 & 68.8 & 41.1 & 74.7 & 74.9 \\ 
PillarNet-18~\cite{pillarnet}$^{\dagger}$  & 65.0 & 70.8 & 78 ms  & 87.4 & 56.7 & 60.9 & 61.8 & 30.4 & 87.2 & 67.4 & 40.3 & 82.1 & 76.0 \\ 
\hline \hline
VoxelNeXt-2D  & 64.1 & 69.8 & 61 ms &  84.8 & 52.7 & 62.3 & 56.2 & 29.5 & 84.5 & 72.5 & 45.7 & 78.8 & 73.7 \\ 
VoxelNeXt  & 64.5 & 70.0  & 66 ms & 84.6 & 53.0 & 64.7 & 55.8 & 28.7 & 85.8 & 73.2 & 45.7 & 79.0 & 74.6 \\ 
VoxelNeXt$^{\dagger}$  & \textbf{66.2} & \textbf{71.4} & - & 85.3 & 55.7 & 66.2 & 57.2 & 29.8 & 86.5 & 75.2 & 48.8 & 80.7 & 76.1 \\ 
\hline
\end{tabular}}
\label{tab:nuscenes-test}
\end{center}
\end{table*}
\begin{table}[t]
\begin{center}
\caption{Performance of nuScenes 3D tracking test split for LIDAR-only methods, without multi-modal extension. $^{\dagger}$ is based on the double-flip 3D object detection results in Tab.~\ref{tab:nuscenes-test}.}
\resizebox{\linewidth}{!}{
\begin{tabular}{|l|c|ccc|}
\hline
{\em Method}                        &   AMOTA    & AMOTP & MOTA & IDS \\ \hline \hline
AB3DMOT~\cite{ab3dmot} & 15.1 & 150.1 & 15.4 & 9027 \\
CenterPoint~\cite{centerpoint} & 63.8 & 55.5 & 53.7 & 760 \\
CBMOT~\cite{cbmot} & 64.9 & 59.2 & 54.5 & 557 \\
OGR3MOT~\cite{ogr3mot} & 65.6 & 62.0 & 55.4 & 288 \\
SimpleTrack~\cite{simpletrack} & 66.8 & 55.0 & 56.6 & 575 \\
UVTR-L~\cite{uvtr} & 67.0 & 55.0 & 56.6 & 774 \\
TransFusion-L~\cite{transfusion} & 68.6 & 52.9 & 57.1 & 893 \\
\hline
VoxelNeXt   & \textbf{69.5} & 56.8 & 58.6 & 785 \\

VoxelNeXt$^{\dagger}$   & \textbf{71.0} & 51.1 & 60.0 & 654 \\
\hline
\end{tabular}}
\label{tab:nuscenee-tracking-test}
\end{center}
\end{table}
\subsection{Ablation Studies}
\noindent
\textbf{Additional Down-sampling Layers}
We ablate the effect of the down-sampling layers in VoxelNeXt.
We extend it to the variants $D_s$. $s$ denotes the number of down-sampling. For example, $D_3$ has the same network strides (3 times) to the base model. Our modification does not change the resolution for the detection head. 
The results of these models are shown in Tab.~\ref{tab:nuscenes-downsamples-ablation}. Without the dense head, $D_3$ suffers from serious performance drop, especially on large objects of {\em Truck} and {\em Bus}. From $D_3$ to $D_5$, performance gradually increases. Additional down-sampling layers compensate for the receptive field. To verify this, we add one more variant, $D_3^{5\times 5\times 5}$, which increases the kernel size of sparse convolutions in all stages to $5\times 5\times 5$. Large kernel improves the performance to some extend but degrades efficiency. 
Thus, we use additional down-samplings as a simple solution.

\begin{table}[t]
\begin{center}
\caption{Performance of nuScenes 3D tracking validation set. All methods listed are LIDAR-only, without multi-modal extension.}
\resizebox{\linewidth}{!}{
\begin{tabular}{|l|c|ccc|}
\hline
{\em Method}                        &   AMOTA    & AMOTP & MOTA & IDS \\ \hline \hline
AB3DMOT~\cite{ab3dmot} & 57.8 & 80.7 & 51.4 & 1275 \\
MPN-Baseline & 59.3 & 83.2 & 51.4 & 1079 \\
CenterPoint~\cite{centerpoint} & 66.5 & 56.7 & 56.2 & 562 \\
CBMOT~\cite{cbmot} & 67.5 & 59.1 & 58.3 & 494 \\
OGR3MOT~\cite{votingforvoting} & 69.3 & 62.7 & 60.2 & 262 \\
SimpleTrack~\cite{simpletrack} & 69.6 & 54.7 & 60.2 & 405 \\
\hline
VoxelNeXt   & \textbf{70.2} & 64.0 & 61.5 & 729 \\
\hline
\end{tabular}
}
\label{tab:nuscenes-tracking-val}
\end{center}
\end{table}
\begin{table*}[t]
\begin{center}
\caption{Performance of 3D object detection results on the Waymo validation split. 
Results with the instance-decreasing trick in the ground-truth sampling~\cite{fsd} is in the appendix.
All models take single-frame data as input without test-time augmentations or ensemble.}
\resizebox{\linewidth}{!}{
\begin{tabular}{|l|c|cc|cc|cc|}
\hline
\multirow{2}{*}{\em Method} & mAP/mAPH  & \multicolumn{2}{c|}{Vehicle} & \multicolumn{2}{c|}{Pedestrian} & \multicolumn{2}{c|}{Cyclist} \\
                                               & L2        & L1 AP/APH           & L2  AP/APH          & L1  AP/APH             & L2     AP/APH         & L1     AP/APH        & L2  AP/APH          \\ \hline \hline
Pillar-OD~\cite{pillar-od}                                        &  -  & 69.8 / -     & - / -    &  72.5 / - &  -       &  -      &  -     \\
VoxSeT~\cite{voxelset}                                        &  -  & 76.0 / -     & 68.2 / -    &  - &  -       &  -      &  -     \\
VoTr-TSD~\cite{voxeltransformer}                                        &  -  & 74.9 / 74.3     & 65.9 / 65.3     & -  &  -       &  -      &  -     \\
SECOND~\cite{second}                                         & 61.0 / 57.2 & 72.3 / 71.7     & 63.9 / 63.3    & 68.7 / 58.2      & 60.7 / 51.3      & 60.6 / 59.3     & 58.3 / 57.0    \\
M3METR~\cite{m3metr}                                        & 61.8 / 58.7 & 75.7 / 75.1     & 66.0 / 66.0    & 65.0 / 56.4      & 56.0 / 48.4      & 65.4 / 64.2     & 62.7 / 61.5    \\
IA-SSD~\cite{ia-ssd}                                       & 62.3 / 58.1 & 70.5 / 69.7     & 61.6 / 61.0    & 69.4 / 58.5      & 60.3 / 50.7      & 67.7 / 65.3     & 65.0 / 62.7    \\
PointPillars~\cite{pointpillars}                                  & 62.8 / 57.8 & 72.1 / 71.5     & 63.6 / 63.1    & 70.6 / 56.7      & 62.8 / 50.3      & 64.4 / 62.3     & 61.9 / 59.9    \\
RangeDet~\cite{rangedet}                                      & 65.0 / 63.2 & 72.9 / 72.3     & 64.0 / 63.6    & 75.9 / 71.9      & 67.6 / 63.9      & 65.7 / 64.4     & 63.3 / 62.1    \\
3D-MAN~\cite{3d-man}                                        &  -  & 74.5 / 74.0 & 67.6 / 67.1 & 71.7 / 67.7 & 62.6 / 59.0   &  -  &  -     \\
LIDAR-RCNN~\cite{lidar-rcnn}                                     & 65.8 / 61.3 & 76.0 / 75.5     & 68.3 / 67.9    & 71.2 / 58.7      & 63.1 / 51.7      & 68.6 / 66.9     & 66.1 / 64.4    \\
PV-RCNN~\cite{pvrcnn}                                        & 66.8 / 63.3 & 77.5 / 76.9     & 69.0 / 68.4    & 75.0 / 65.6      & 66.0 / 57.6      & 67.8 / 66.4     & 65.4 / 64.0    \\
Part-A2-Net~\cite{part-a2}                                    & 66.9 / 63.8 & 77.1 / 76.5     & 68.5 / 68.0    & 75.2 / 66.9      & 66.2 / 58.6      & 68.6 / 67.4     & 66.1 / 64.9    \\
SST~\cite{single-stride-transformer}                                            & 67.8 / 64.6 & 74.2 / 73.8     & 65.5 / 65.1    & 78.7 / 69.6      & 70.0 / 61.7      & 70.7 / 69.6     & 68.0 / 66.9    \\
PV-RCNN++~\cite{pvrcnn++}                                      & 68.4 / 64.9 & 78.8 / 78.2     & 70.3 / 69.7    & 76.7 / 67.2      & 68.5 / 59.7      & 69.0 / 67.6     & 66.5 / 65.2    \\
CenterPoint~\cite{centerpoint}                                    & 69.8 / 67.6 & 76.6 / 76.0     & 68.9 / 68.4    & 79.0 / 73.4      & 71.0 / 65.8      & 72.1 / 71.0     & 69.5 / 68.5    \\
AFDetV2~\cite{afdetv2}                                        & 71.0 / 68.8 & 77.6 / 77.1     & 69.7 / 69.2    & 80.2 / 74.6      & 72.2 / 67.0      & 73.7 / 72.7     & 71.0 / 70.1    \\
PillarNet-34~\cite{pillarnet}                                   & 71.0 / 68.5 & {79.1} / {78.6}     & {70.9} / {70.5}    & 80.6 / 74.0      & 72.3 / 66.2      & 72.3 / 71.2     & 69.7 / 68.7    \\
SWFormer~\cite{swformer}                                       & -         & 77.8 / 77.3     & 69.2 / 68.8    & 80.9 / 72.7      & 72.5 / 64.9      & -             & -            \\ 
FSD$_{spconv}$~\cite{fsd}                                      & 71.9 / 69.7 & 77.8 / 77.3     & 68.9 / 68.5    & 81.9 / 76.4      & 73.2 / 68.0      & 76.5 / 75.2     & 73.8 / 72.5    \\
\hline
VoxelNeXt-2D            &    70.9 / 68.2       & 77.9 / 77.5     & 69.7 / 69.2    & 80.2 / 73.5      & 72.2 / 65.9      & 73.3 / 72.2     & 70.7 / 69.6    \\
VoxelNeXt$_{K3}$    &    \textbf{72.2} / \textbf{70.1}       & 78.2 / 77.7     & 69.9 / 69.4    & 81.5 / 76.3      & 73.5 / 68.6      & 76.1 / 74.9     & 73.3 / 72.2    \\
\hline
\end{tabular}
}
\label{tab:waymo-val}
\end{center}
\end{table*}
\begin{table*}[t]
\begin{center}
\caption{Performance of 3D object detection results Argoverse2 dataset.}
\vspace{-5pt}
\resizebox{\linewidth}{!}{
\begin{tabular}{|l|c|cccccccccccccccccccc|}
\hline
Methods & mAP & Veh. & Bus & Ped. & Stop. & Box. & Boll. & C-B. & M.-list & MPC. & M.-cycle & Bicycle & A-B. & School. & Truck. & C-C. & V-T. & Sign & Large. & Str. & Bic.-list \\ \hline
CenterPoint~\cite{centerpoint} & 22.0 & 67.6 & 38.9 & 46.5 & 16.9 & 37.4 & 40.1 & 32.2 & 28.6 & 27.4 & 33.4 & 24.5 & 8.7 & 25.8 & \textbf{22.6} & 29.5 & 22.4 & 6.3 & 3.9 & 0.5 & 20.1 \\
FSD & 28.2 & 68.1 & \textbf{40.9} & 59.0 & 29.0 & 38.5 & 41.8 & 42.6 & 39.7 & 26.2 & \textbf{49.0} & 38.6 & \textbf{20.4} & \textbf{30.5} & 14.8 & 41.2 & \textbf{26.9} & 11.9 & 5.9 & 13.8 & 33.4 \\ \hline
VoxelNeXt & 30.0 & 71.7 & 39.2 & 63.1 & 39.2 & 40.0 & 52.5 & 63.7 & 42.2 & 34.9 & 42.7 & 40.1 & 20.1 & 25.2 & 16.9 & \textbf{45.7} & 22.3 & \textbf{15.8} & 5.9 & 9.8 & \textbf{33.5} \\
VoxelNeXt$_{K3}$ & \textbf{30.7} & \textbf{72.7} & 38.8 & \textbf{63.2} & \textbf{40.2} & \textbf{40.1} & \textbf{53.9} & \textbf{64.9} & \textbf{44.7} & \textbf{39.4} & 42.4 & \textbf{40.6} & 20.1 & 25.2 & 19.9 & 44.9 & 20.9 & {14.9} & \textbf{6.8} & \textbf{15.7} & 32.4 \\
\hline
\end{tabular}}
\label{tab:argo2-allclasses}
\end{center}
\end{table*}
\vspace{0.5em}
\noindent
\textbf{Spatially Voxel Pruning}
VoxelNeXt gradually drops redundant voxels according to feature magnitude. We ablate this setting in Tab.~\ref{tab:nuscenes-sprs-ratio-ablation}. We control the drop ratio from 0.1 to 0.9 with an interval of 0.2.
The performance hardly decays when the ratio is not greater than 0.5. Thus, we set the drop ratio to 0.5 as a default setting in our experiments. We also ablate the stages of voxel pruning in Tab.~\ref{tab:voxel-pruning-layers}. We use it on the first 3 stages by default.

\vspace{0.5em}
\noindent
\textbf{Sparse Height Compression}
We make ablations on the sparse CNN types of 2D and 3D, in the backbone and head of VoxelNeXt, in Tab.~\ref{tab:ablation-sparse-bev-compression}. The naive design is that both the backbone and head apply 3D sparse CNN, which results in high latency. With the sparse height compression, we combine the 3D backbone and 2D sparse prediction head. It achieves much better efficiency with decent performance. We use it as a default setting of VoxelNeXt. When we use 2D sparse CNN as the backbone network, it has the same layer number and double channels as the 3D one. It achieves the best efficiency, and yet suffers a bit of performance drop. We name it VoxelNeXt-2D for its high efficiency.

\vspace{0.5em}
\noindent
\textbf{Layer Type in Sparse Prediction Head}
We ablate the effect using fully-connected layers or submanifold sparse convolutions to predict boxes in the sparse head, as shown in Tab.~\ref{tab:head-sparse-conv}. The fully-connected~(FC) head has inferior performance to the $3\times 3$ sparse convolution counterpart, but more efficient. We denote the latter with $K3$ in VoxelNeXt. 

\vspace{0.5em}
\noindent
\textbf{Relative Positions between Voxels and Predicted Boxes}
In VoxelNeXt, voxels for box prediction are not required to be inside the boxes, not to mention centers, as in Tab.~\ref{tab:ratio-voxel-inside}. We count the relative of voxels that are inside the 3D bounding boxes they generate. We split voxels into 3 area types of {\em near center}, {\em near boundary}, and {\em outside box}, according to their relative positions to boxes. On average, most boxes are predicted from voxels inside, maybe not near centers. Statistically, only a few boxes (less than 10\% in total) are predicted based on the voxels near object centers. This finding shows that boundary voxels are also qualified for prediction, while object centers are not always necessary.

Another observation is that there are large gaps between the ratios of different classes. For {\em Car} and {\em Trailer}, most boxes are predicted on inside voxels. In contrast, for {\em Truck}, {\em Traffic Cone}, and {\em Pedestrian}, about half of the boxes are predicted from outside voxels. We illustrate example pairs in Fig.~\ref{fig:voxels-and-points}. As objects in different classes vary in size and spatial sparsity, predicting upon voxels complies with data distribution, rather than proxies like anchors or centers.

\vspace{0.5em}
\noindent
\textbf{Comparison to CenterPoint in Error Analysis}
We compare VoxelNeXt to the representative dense-head method
CenterPoint~\cite{centerpoint} in Tab.~\ref{tab:nuscenes-error-analysis}. Training on 1/4 nuScenes training set and evaluating on the full validation split, VoxelNeXt achieves 0.9\% mAP and 1.0\% NDS improvement. In further analysis, CenterPoint and VoxelNeXt shares comparable errors in location, size, and velocity. However, there are large gaps in other error types, especially in orientation. Notably, VoxelNext has 4.9\% less orientation error than CenterPoint. We suppose that this results from that sparse voxel features might be more sensitive to orientation difference.

\vspace{0.5em}
\noindent
\textbf{Efficiency Statistics of Backbone}
We count the efficiency-related statistics of our sparse CNN backbone network in Tab.~\ref{tab:nuscenes-efficiency-stages}. As features in the last 3 stages are summed up for height compression, they share the same channel number 128. Due to the high down-sampling ratios in Stages 5-6, their voxel numbers are much smaller compared to previous stages. Consequently, the computation cost introduced in Stages 5-6 is limited to 6.1G and 2.8G FLOPs in 6 and 3 ms. It is no more than $1/3$ of the overall backbone network, and yet makes notable effects on performance enhancement.

\vspace{0.5em}
\noindent
\textbf{Sparse Max Pooling}
We ablate the effect of sparse max pooling and NMS in Tab.~\ref{tab:nuscenes-maxpool-ablation}. Compared to the commonly used NMS, max-pool presents comparable mAP, 56.0\% v.s. 56.2\%. VoxelNeXt is flexible to works either with NMS or sparse max pooling. Max-pool is an elegant solution and avoids some unnecessary computation on predictions.

\vspace{0.5em}
\noindent
\textbf{Voxel Association for 3D Tracking}
Tab.~\ref{tab:nuscenes-voxel-tracking} shows the ablation of 3D tracking on nuScenes validation. In addition to tracking predicted box centers, we also include the voxels that predict boxes for matching. Voxel association introduces notable improvement of 1.1\% AMOTA. 

\subsection{Main Results}
\noindent
\textbf{3D Object Detection}
In Tab.~\ref{tab:nuscenes-test}, we evaluate our detection models on the test split and compare them with other LIDAR-based methods on nuScenes test set.
Results denoted as $\dagger$~\cite{vista, uvtr, pillarnet} are reported with the double-flip testing augmentation~\cite{centerpoint}. Both lines of results are better than previous ones. 
We compare VoxelNeXt with other 3D object detectors on the Waymo validation split in Tab.~\ref{tab:waymo-val} and on Argoverse2~\cite{argo2} in Tab.~\ref{tab:argo2-allclasses}. 
We present latency comparison in Tab.~\ref{tab:nuscenes-test} and Fig.~\ref{fig:argo2-efficiency-comparison}.
VoxelNeXt achieves leading performance among these methods with high efficiency. 

\vspace{0.1em}
\noindent
\textbf{3D Multi-object Tracking}
In Tab.~\ref{tab:nuscenee-tracking-test} and Tab.~\ref{tab:nuscenes-tracking-val}, we compare VoxelNeXt's tracking performance with other methods in the nuScenes test and validation splits. VoxelNeXt achieves the best AMOTA among all LIDAR-based methods. In addition, when combined with the double-flip testing results in Tab.~\ref{tab:nuscenes-test}, denoted as $^{\dagger}$ in Tab.~\ref{tab:nuscenee-tracking-test}, VoxelNeXt further achieves 71.0\% AMOTA and ranking 1$^{st}$ on the nuScenes 3D LIDAR tracking benchmark.

\section{Conclusion and Discussion}
\label{sec:conclusion}
In this paper, we have presented a fully sparse and voxel-based framework for 3D object detection and tracking. It is with simple techniques, run fast with no much extra cost, and works in an elegant manner without NMS post-processing. For the first time, we show that direct voxel-based prediction is feasible and effective. Thus rule-based schemes, {\em e.g.}, anchors or centers, and dense heads become unnecessary in ours. VoxelNeXt presents promising results on large-scale datasets, including nuScenes~\cite{nuscenes}, Waymo~\cite{waymo}, and Argoverse2~\cite{argo2}. With high efficiency, it achieves leading performance on 3D object detection and ranks 1$_{st}$ on nuScenes 3D tracking LIDAR benchmark.

\vspace{0.35em}
\noindent
\textbf{Limitations}
A gap exists between theoretical FLOPs and actual inference speed.
VoxelNeXt has a much small 38.7G FLOPs, compared to 186.6G of CenterPoint~\cite{centerpoint}. The actual latency reduction is clear but not so large as FLOPs in Tab.~\ref{tab:computation-comparison}, as it highly depends on implementation and devices.

{\small
\bibliographystyle{ieee_fullname}
\bibliography{egbib}
}

\appendix
\captionsetup[table]{labelformat={default},labelsep=period,name={Table A -}}
\captionsetup[figure]{labelformat={default},labelsep=period,name={Figure A -}}

\section*{Appendix}
In this appendix, we first introduce implementation details in Sec.~\ref{sec:implementation-details}. We then include additional experimental results in Sec.~\ref{sec:experiments-results}. We also provide more visualizations and discussions in Sec.~\ref{sec:visualizations} and Sec.~\ref{sec:discussion}. 

\section{Implementation Details}
\label{sec:implementation-details}

\noindent
\textbf{nuScenes}~\cite{nuscenes} has 1,000 drive sequences, split into 700, 150, and 150 sequences for training, validation, and testing. nuScenes is collected by a 32-beam synced LIDAR and 6 cameras. The annotations include 10 classes. In the ablation study, detection models are trained on 1/4 training data and evaluated on the full validation set. 

\vspace{0.5em}
\noindent
\textbf{Waymo}~\cite{waymo} is a large-scale public autonomous driving dataset, which contains 1,150 sequences in total, with 798 for training, and 202 for validation. It was collected by one long-range LiDAR sensor at 75 meters and four near-range sensors. 

\vspace{0.5em}
\noindent
\textbf{Argoverse2}~\cite{argo2} has 1000 sequences, including 700 for training, 150 for validation. The perception range is 200 radius meters, covering area of 400m × 400m. We follow FSD~\cite{fsd} for data processing.

\vspace{0.5em}
\noindent
\textbf{Voxelization}

For nuScenes~\cite{nuscenes} dataset, point clouds are clipped in [-54m, 54m] for {\em X} or {\em Y} axis, and [-5m, 3m] for {\em Z} axis. Voxel size is (0.075m, 0.075m, 0.2m) by default. For VoxelNeXt-2D, the voxel size along {\em Z} axis is 8m. 

For Waymo~\cite{waymo} dataset, point clouds are clipped into [-75.2m, 75.2m] {\em X} or {\em Y} axis, and [-2m, 4m] for {\em Z} axis. Voxel size is (0.1m, 0.1m, 0.15m) by default.  For VoxelNeXt-2D, the voxel size along {\em Z} axis is 6m. 

For Argoverse2~\cite{argo2} dataset, we use (0.1m, 0.1m, 0.2m) as voxel size. The perception range is [-200m, 200m] for {\em X} or {\em Y} axis. The range for {\em Z} is [-20m, 20m].

\vspace{0.5em}
\noindent
\textbf{Data Augmentations}

For nuScenes dataset, random flipping, global scaling, global rotation, GT sampling~\cite{second}, and  translation augmentations are used. Flipping is randomly conducted along {\em X} and {\em Y} axes. Rotation angle is randomly picked between -45$^{\mathrm{o}}$ and 45$^{\mathrm{o}}$. Global scaling is conducted by a factor sampled between 0.9 and 1.1. The translation noise factors are sampled between 0 and 0.5. Only for test submission models, GT sampling is removed in the last 5 training epochs~\cite{uvtr}. 

For Waymo dataset, data augmentations also include random flipping, global scaling, global rotation, and ground-truth (GT) sampling~\cite{second}. These settings are similar to those of nuScenes dataset and follow baseline methods~\cite{centerpoint,pvrcnn}.

For Argoverse2 dataset, we use similar data augmentation to nuScenes and Waymo, except that we do not use ground-truth sampling.

\begin{figure}[t]
\begin{center}
   \includegraphics[width=\linewidth]{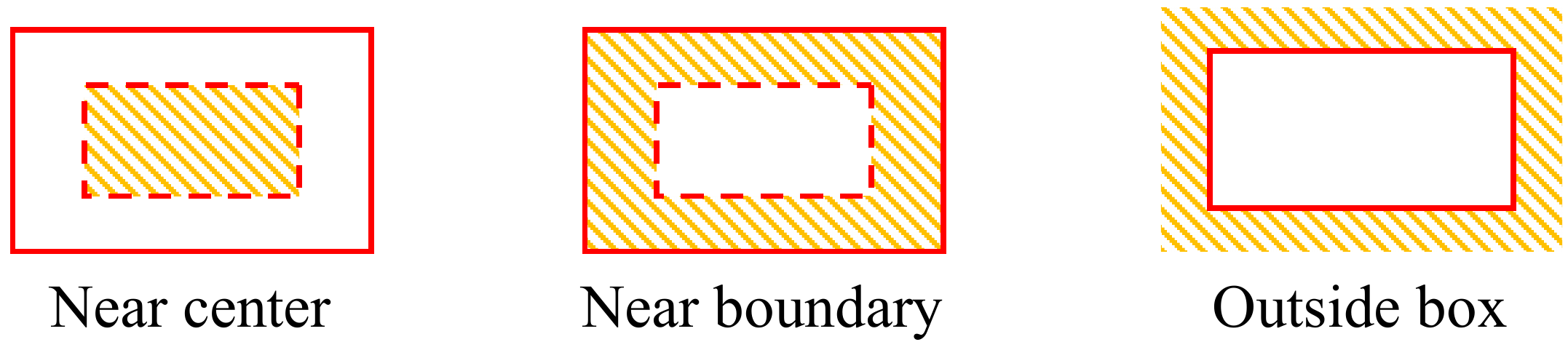}
   \caption{The relative positions of query voxel to the predicted boxes, {\em e.g.}, {\em near center}, {\em near boundary}, {\em outside box}, corresponding to Tab.~\textcolor{red}{7} in the paper.}
   \label{fig:boxe-areas}
\end{center}
\end{figure}
\vspace{0.5em}
\noindent
\textbf{Training Hyper-parameters}

For nuScenes dataset, models are trained for 20 epochs with batch size 16. They are optimized with Adam~\cite{adam}. Learning rate is initially 1e-3 and decays to 1e-4 in a cosine annealing. Weight decay is 0.01. Gradients are clipped by norm 35. These settings follow CenterPoint~\cite{centerpoint}. 

For Waymo dataset, models are trained for 12 epochs by default. Batch size is set as 16. Learning rate is initialized as 3e-3. They are also optimized with Adam~\cite{adam}. 

For Argoverse2 dataset, we use similar settings to Waymo, except that only 6 epochs for training is enough.

\begin{table*}[t]
\begin{center}
\caption{Comparison on the nuScenes validation split. This table presents detailed performance for Tab.~\textcolor{red}{1} in the paper.}
\begin{tabular}{|l|cc|cccccccccc|}
\hline
                   {\em Method}                 & mAP   & NDS & Car                  & Truck                & Bus                  & Trailer              & C.V.                   & Ped                  & Mot                  & Byc                  & T.C.                   & Bar                   \\ \hline
SECOND~\cite{second}   &  50.6 & 62.3 & 81.8 & 51.7 & 66.9 & 37.3 & 15.0 & 77.7 & 42.5 & 17.5 & 57.4 & 59.2 \\
CenterPoint~\cite{centerpoint}  & 58.6 & 66.2 & 85.0 & 58.2 & 69.5 & 35.7 & 15.5 & 85.3 & 58.8 & 40.9 & 70.0 & 67.1 \\
VoxelNeXt    & 60.0 & 67.1 & 85.6 & 58.4 & 71.6 & 38.6 & 17.9 & 85.4 & 59.7 & 43.4 & 70.8 & 68.1 \\  \hline
\end{tabular}
\label{tab:nuscenes-validation}
\end{center}
\end{table*}
\begin{table}[t]
\begin{center}
\caption{Effects of the feature levels for prediction.}
\begin{tabular}{|c|cc|}
\hline
Head resolution   &   mAP    & NDS  \\ \hline
 8 & \textbf{56.2} & \textbf{64.3} \\
16 & 52.5 & 60.7 \\
32 & 49.0 & 57.9 \\
\{8, 16, 32\} & 55.7 & 63.7 \\
\{2, 4, 8, 16, 32\} & 53.9 & 62.2 \\
\hline
\end{tabular}
\label{tab:nuscenes-output-downsamples-ablation}
\end{center}
\end{table}
\begin{table}[t]
\begin{center}
\caption{Gap between VoxelNeXt-2D and VoxelNet. mAP on nuScenes validation with different amounts of training data.}
\begin{tabular}{|l|ccc|}
\hline
{\em Method} & 1/4  & 1/2  & full \\ \hline
VoxelNeXt-2D & 53.4  & 56.0  & 58.7 \\ \hline
VoxelNeXt    & 56.2 & 58.2 & 60.0 \\ \hline
\end{tabular}
\label{tab:analysis-data-amount}
\end{center}
\end{table}

\begin{figure*}[t]
\begin{center}
   \includegraphics[width=\linewidth]{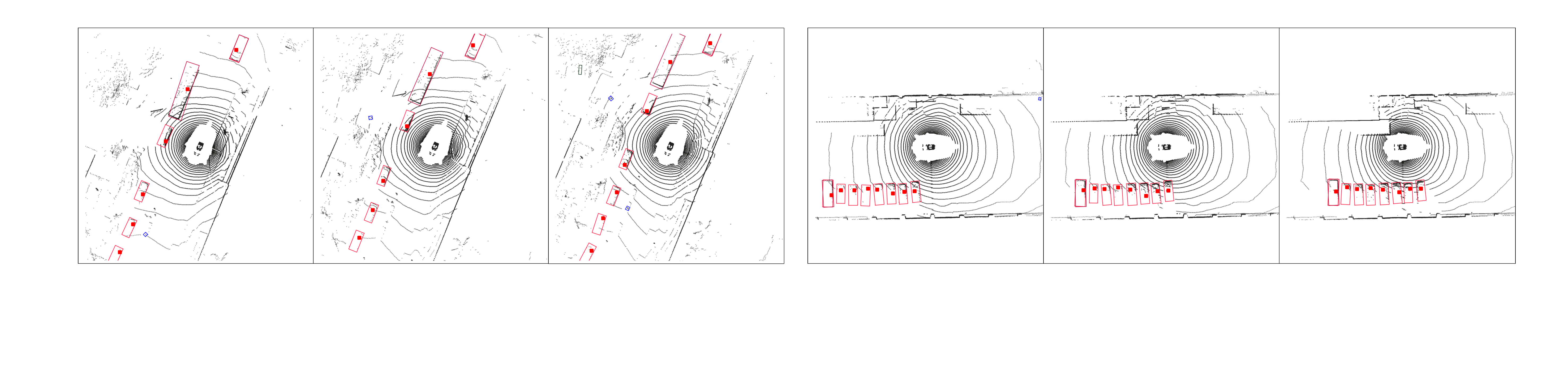}
   \caption{Detections of adjacent frames. We visualize predicted boxes and the corresponding query voxels, which are enlarged as red squares. This figure is best viewed by zoom-in.}
   \label{fig:frames-box-query-voxels}
\end{center}
\end{figure*}

\vspace{0.5em}
\noindent
\textbf{Network Structures}

We develop our VoxelNeXt network upon the widely-used residual sparse convolutional block~\cite{centerpoint,pvrcnn,voxel-rcnn}. We use 2D sparse convolutions in its variant of VoxelNeXt-2D. For voxel selection and box regression, we use fully-connected layer or kernel-size-3 submanifold sparse convolutions~\cite{submanifold-sparse-conv-v2} for prediction. The former convolution has 128 channels in VoxelNeXt-2D and 64 in 3D networks. Training schedules and hyper-parameters follow prior works~\cite{pvrcnn,centerpoint}. 

The backbone network of  VoxelNeXt has 6 stages. The channels for these stages are \{16, 32, 64, 128, 128, 128\} by default. There are 2 residual submanifold sparse convolutional blocks~\cite{submanifold-sparse-conv-v2} in each stage. The sparse head predicts outputs by $3\times 3$ submainfold sparse convolutions. Following CenterPoint~\cite{centerpoint}, the prediction layers are only shared among similar classes on nuScenes and Argoverse2 and shared among all classes on Waymo. The kernel sizes for the sparse max pooling layer varies in different heads, because the size of objects varies in different classes.

\section{Experimental results}
\label{sec:experiments-results}
\noindent
\textbf{Performance on nuScenes Validation}$\;$
We provide the performance of VoxelNeXt on nuScenes {\em val} in Tab.~A~-~\ref{tab:nuscenes-validation}.

\vspace{0.5em}
\noindent
\textbf{Gaps between VoxelNeXt and VoxelNeXt-2D}$\;$
We analyze the gaps between VoxelNeXt and VoxelNeXt-2D on different amounts of training data in Tab.~A~-~\ref{tab:analysis-data-amount}. These models are trained on 1/4, 1/2, and full nuScenes training set, respectively, and evaluated on the full validation set. It shows that The gap is large on the 1/4 training data, while the gaps gradually narrow as the data amount grows. Overall, the 3D network can obtain much better performance than its 2D counterpart at a small amount of data. Meanwhile, VoxelNeXt-2D has potential on large data amount.

\vspace{0.5em}
\noindent
\textbf{Resolution of Sparse Head}
We make an ablation study on the resolution of prediction head in Tab.~A~-~\ref{tab:nuscenes-output-downsamples-ablation}. The performance decreases as the head resolution increases from the default setting of 8 to 32. In addition, we also evaluate the multi-head design of \{8, 16, 32\} and \{2, 4, 8, 16, 32\}, where results are combined from the multiple heads with various resolutions. These multi-head models present no better results than the single-resolution 8 network.

\begin{table}[t]
\begin{center}
\caption{Results on Vehicle detection on Waymo. $^*$ means decreasing the number of pasted instances in the ground-truth sampling augmentation and increase training epochs by 6 epochs~\cite{fsd}.}
\begin{tabular}{|l|ll|}
\hline
 {\em Method} & L1 AP/APH           & L2  AP/APH          \\ \hline 
VoxelNeXt    &  78.2 / 77.7     & 69.9 / 69.4  \\ 
VoxelNeXt$^*$    &   79.1 / 79.0 &  70.8 / 70.5 \\ \hline
\end{tabular}
\label{tab:waymo-vehicle}
\end{center}
\end{table}

\vspace{0.5em}
\noindent
\textbf{Performance on Waymo vehicle detection}$\;$
In Tab.~A~-~\ref{tab:waymo-vehicle}, we follow FSD~\cite{fsd} to decrease the number of pasted instances in the ground-truth sampling augmentation and increase training epochs by 6 epochs. This trick leads to better results upon VoxelNeXt on the Waymo object detection.

\section{Visualizations}
\label{sec:visualizations}
We visualize the results of adjacent frames in Fig.~A~-~\ref{fig:frames-box-query-voxels}. The corresponding query voxels are depicted as red squares. 

\section{Discussions}
\label{sec:discussion}
\noindent
\textbf{Point-based Detectors}$\;$
Point-based 3D object detectors~\cite{votenet,point-rcnn,3dssd,ia-ssd} are fully sparse by their very nature. Point R-CNN~\cite{point-rcnn} is a pioneer work and presents decent performance on KITTI~\cite{kitti}. Methods of SSD series~\cite{3dssd}, including 3DSSD~\cite{3dssd,ia-ssd}, inherit the point-based tradition and accelerate the methods with simplified pipelines. VoteNet~\cite{votenet} is based on center voting and studies indoor 3D object detection. However, point-based detectors are usually used in scenes with limited points. The neighborhood query operation is still unaffordable in large-scale benchmarks~\cite{nuscenes,waymo}, which are dominated by voxel-based detectors~\cite{pvrcnn,centerpoint}.

\vspace{0.5em}
\noindent
\textbf{Boarder Impacts}$\;$
VoxelNeXt replies on 3D data and its spatially sparse distribution. It might reflect biases in data collection, including the ones of negative societal impacts.

\end{document}